\documentclass[letterpaper]{article} % DO NOT CHANGE THIS
\usepackage[submission]{aaai25}  % DO NOT CHANGE THIS
\usepackage{times}  % DO NOT CHANGE THIS
\usepackage{helvet}  % DO NOT CHANGE THIS
\usepackage{courier}  % DO NOT CHANGE THIS
\usepackage[hyphens]{url}  % DO NOT CHANGE THIS
\usepackage{graphicx} % DO NOT CHANGE THIS
\usepackage{natbib}  % DO NOT CHANGE THIS AND DO NOT ADD ANY OPTIONS TO IT
\usepackage{caption} % DO NOT CHANGE THIS AND DO NOT ADD ANY OPTIONS TO IT
\usepackage{algorithm}
\usepackage{algorithmic}

\usepackage{amsmath}

\usepackage{multirow}
\usepackage{colortbl}
\usepackage{booktabs}
\usepackage{colortbl}

\usepackage{makecell}
\usepackage{pifont}
\usepackage{newfloat}
\usepackage{listings}
\DeclareCaptionStyle{ruled}{labelfont=normalfont,labelsep=colon,strut=off} % DO NOT CHANGE THIS
\floatstyle{ruled}
\newfloat{listing}{tb}{lst}{}
\floatname{listing}{Listing}
\title{Inf-MLLM: Efficient Streaming Inference of Multimodal Large Language Models on a Single GPU}
\author{
    Zhenyu Ning\textsuperscript{\rm 1},
    Jieru Zhao\textsuperscript{\rm 1}\thanks{Corresponding author: Jieru Zhao (zhao-jieru@sjtu.edu.cn)},
    Qihao Jin\textsuperscript{\rm 2},
    Wenchao Ding\textsuperscript{\rm 2},
    Minyi Guo\textsuperscript{\rm 1}
}
\affiliations{
    \textsuperscript{\rm 1}Department of Computer Science and Engineering, Shanghai Jiao Tong University\\
    \textsuperscript{\rm 2}Academy for Engineering and Technology, Fudan University\\
    \{2336631036, zhao-jieru\}@sjtu.edu.cn,
    \{20307130014, dingwenchao\}@fudan.edu.cn, guo-my@cs.sjtu.edu.cn
}

\usepackage{bibentry}
\begin{document}

\maketitle

\begin{abstract}
% We propose Inf-MLLM, an efficient streaming inference framework enabling MLLMs to model infinite context on a single GPU, based on our special attention pattern and KV cache mechanism. 
Multimodal Large Language Models (MLLMs) are distinguished by their multimodal comprehensive ability and widely used in many real-world applications including GPT-4o, autonomous driving and robotics. Despite their impressive performance, the multimodal inputs always incur long context. The inference under long context requires caching massive Key and Value states (KV cache) of previous tokens, which introduces high latency and excessive memory consumption. Due to this reason, it is challenging to deploy streaming inference of MLLMs on edge devices, which largely constrains the power and usage of MLLMs in real-world applications. In this paper, we introduce Inf-MLLM, an efficient \underline{inf}erence framework for MLLMs, which enable streaming inference of MLLM on a single GPU with \underline{inf}inite context. Inf-MLLM is based on our key observation of the attention pattern in both LLMs and MLLMs called ``attention saddles". Thanks to the newly discovered attention pattern, Inf-MLLM maintains a size-constrained KV cache by dynamically caching recent tokens and relevant tokens. Furthermore, Inf-MLLM proposes attention bias, a novel approach to enable MLLMs to capture long-term dependency. We show that Inf-MLLM enables multiple LLMs and MLLMs to achieve stable performance over 4M-token long texts and multi-round conversations with 1-hour-long videos on a single GPU. In addition, Inf-MLLM exhibits superior streaming reasoning quality than existing methods such as StreamingLLM and 2x speedup than H2O. 

% Inf-MLLM leverages the observation that a small specific subset of tokens(\textless 10\%) composed of recent tokens and relevant tokens contribute most of the attention weight(\textgreater 90\%), and only retains these specific tokens in the KV cache to reduce its memory footprint. To support streaming inference on infinite context, Inf-MLLM proposes a novel KV cache eviction mechanism to identify the crucial tokens while making the model always focus on new request. 
% We validate that this approach achieves better performance than previous works on over 1M tokens with limited memory usage. 

\end{abstract}

% Uncomment the following to link to your code, datasets, an extended version or similar.
%
% \begin{links}
%     \link{Code}{https://aaai.org/example/code}
%     \link{Datasets}{https://aaai.org/example/datasets}
%     \link{Extended version}{https://aaai.org/example/extended-version}
% \end{links}

\section{Introduction}
\begin{figure}[tp]
\centering
\includegraphics[width=0.99\columnwidth]{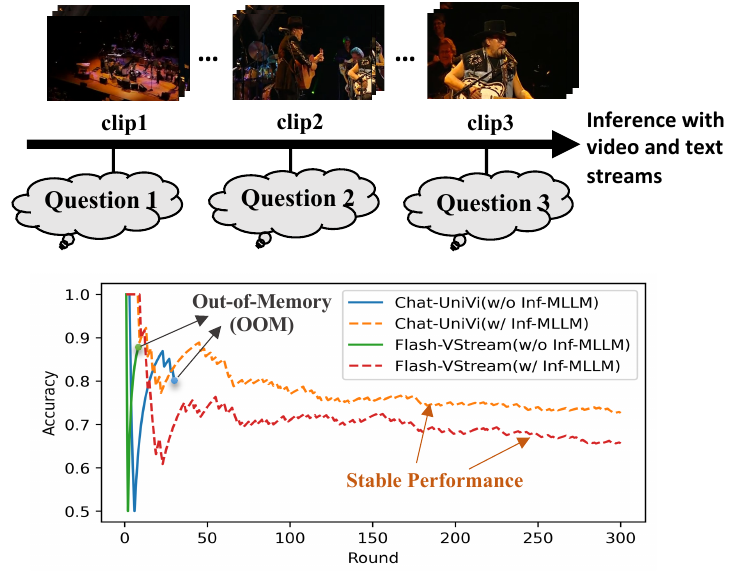}
\caption{Illustration of the streaming inference process. The bottom figure shows that Inf-MLLM facilitates existing MLLMs to handle streams of texts and videos without OOM while maintaining high-quality token generation.}
\label{fig:intro}
\end{figure}

% Large Language Models (LLMs) are exhibiting remarkable advancements in reasoning and generalization abilities in Natural Language Processing (NLP) with the scaling up of data, and gradually becoming the mainstream solution for almost all of the AI tasks \cite{few-shot, llama2}. To extend LLMs to support more complicated real-world problems, 
Multimodal Large Language Models (MLLMs) \cite{gao2024sphinx, alayrac2022flamingo, li2022blip, team2023gemini} have been introduced to empower Large Language Models (LLMs) with new capabilities to process information of different modalities such as image, video, audio, etc \cite{visual-tuning}.
%, through aligning the multimodal inputs with the text . 
%For instance, MLLMs facilitate numerous complex applications related to video, including video captioning \cite{video_captioner}, movie audio descriptions \cite{autoad2}, video-text retrieval \cite{lavila} \cite{mPLUG-video} and video question-answering \cite{llama_adapter, video_llama}. 
% These applications exemplify the multimodal reasoning capabilities of MLLMs.
Video applications, which typically involve lengthy sequence lengths, exemplify the remarkable multimodal reasoning capabilities of MLLMs. However, they also result in significant memory consumption and a decline in model performance when the context length exceeds a certain threshold. These issues are exacerbated in scenarios of \textbf{streaming} inference, as shown in Fig. \ref{fig:intro}, where multimodal inputs are streamed in and MLLMs have to deal with long context or multi-round conversions continuously. 
% To ensure inference efficiency, it is crucial to empower MLLMs with the ability to process long context or even perform multi-round conversations efficiently, which we refer to \textbf{streaming} inference, on a resource-constrained device like a single GPU.

Efficient streaming inference is crucial for many real-world applications. For instance, OpenAI's new flagship model, GPT-4o \cite{gpt4o}, demonstrates efficient inference for video, audio, and text streams. However, it is not open-source and does not facilitate streaming inference on a \textit{local} device without cloud access. Accessing a cloud-scale model through APIs can raise privacy concerns and incur additional costs. For other edge applications like robotics, cloud-scale model is not always accessible, making streaming inference on edge important. 
However, it is challenging to deploy MLLM in such real-world edge applications due to limited memory budget and high efficiency requirement. 

In this paper, we focus on efficient streaming inference of MLLMs on a single GPU and summarize the challenges in four different aspects as follows.  

\noindent \underline{\textit{C1: Quadratic computation complexity}}: The computation complexity of attention is quadratic to the KV cache size, and retrieving KV states incurs additional memory accesses \cite{dao2022flashattention, sukhbaatar2019adaptive, performers}. As the sequence length grows, the decoding speed will decrease to an intolerable extent, especially for multi-round conversation and long video understanding.

\noindent \underline{\textit{C2: Memory consumption}}: For MLLMs, a large KV cache is maintained to avoid re-computation during inference, which scales linearly with the sequence length. This can result in high memory consumption \cite{pope2023efficiently}. The problem is even more severe for multimodal inputs which are transformed into a large number of tokens. For example, a several-minute-long video can be converted into thousands of tokens \cite{chat_univi, llama-vid}.

\noindent \underline{\textit{C3: Context length limitation}}: Since most MLLMs are fine-tuned with pre-trained LLMs, they are constrained by the context window. When sequence length exceeds the length of the training text, the performance degrades soon, which is unacceptable in real-world applications. Therefore, the techniques of length extrapolation are required to deal with overlong inputs \cite{alibi,rope}.
% It's very challenging for MLLMs to perform long sequence reasoning with high accuracy for the paucity of high-quality multimodal datasets \cite{hudson2019gqa, maaz2023videochatgpt, li2023otter} and the exorbitant cost of model fine-tuning \cite{yu2024rlhfv}. MLLMs are constrained by the attention window during pre-training or fine-tuning, which is a consequence of the limited multimodal training data. The majority of existing video understanding benchmarks \cite{msrvtt, msvd, li2024mvbench, li2023seedbench} are made up of short videos of a relatively short duration, typically a few seconds.

\noindent \underline{\textit{C4: Long-term memory}}: The ability to capture long-term dependency is critical for streaming inference of MLLMs. 
% For example, it is necessary to capture long-term dependency between video multi-round conversations with video clips. 
However, it is hard to achieve due to the lack of high-quality multimodal datasets \cite{hudson2019gqa, maaz2023videochatgpt, li2023otter} and cost of fine-tuning \cite{yu2024rlhfv}. Existing video QA datasets \cite{msrvtt, msvd, li2024mvbench, li2023seedbench} contain several-second-long videos and short conversations, which cannot enhance the long-term reasoning capability of MLLMs during finetuning.

Prior studies, such as window attention \cite{beltagy2020longformer, jiang2023mistral, liu2022swin, dong2022cswin}, H2O \cite{zhang2024h2o}, Keyformer \cite{adnan2024keyformer} and StreamingLLM \cite{streamingllm}, improve the inference performance of LLMs, but none of them can handle all the challenges simultaneously, especially for the streaming inference of MLLMs. Although H2O and StreamingLLM enable LLMs to work on super long texts, they either achieve unstable perplexity on long texts or fail on tasks that demand long-term memory. Details can be seen in Section \ref{sec:related works}.
% and extensive data dependency. 
% such as long document question-answering and long video question-answering. 
Moreover, existing methods focus on pure text inputs and cannot be applied to MLLMs with multimodal inputs directly. 

To this end, we propose Inf-MLLM, an innovative inference framework that enables efficient and high-quality streaming inference of MLLMs on a single GPU with infinite text and video streams as input. 
%Inf-MLLM can handle all the challenges (C1-C4). 
We propose an effective KV cache eviction mechanism based on our key observation that there exist critical tokens with high attention scores, like a series of saddle points in non-linear curves.
%, which are \textit{most relevant} to the generation of the current token and should be cached to ensure good performance on streams of videos and texts. 
Borrowing the concept of saddle points in mathematics, we call these tokens as \textbf{attention saddles}. By caching the most relevant tokens and evicting less important KV states of irrelevant tokens, Inf-MLLM improves decoding speed (C1), reduces memory usage (C2), and enables existing MLLMs to support much longer sequence length than its original maximum context length without re-training and fine-tuning (C3). 
%Previous methods like ${\rm H_2O}$ \cite{zhang2024h2o} also try to identify relevant tokens 
We observe that simply aggregating attention scores for each token causes the summation of scores leaning towards earlier tokens in the sequence, making it hard to select real relevant tokens.
% However, this makes larger attention scores accumulate towards earlier tokens in the sequence while important latter tokens are ignored, leading to model collapse on long context reasoning. 
To solve this issue, 
% that larger attention scores accumulate towards earlier tokens which causes model collapse on long context reasoning, 
we further introduce \textbf{attention bias} to ensure that the KV cache continuously evicts earlier tokens and accommodates new attention saddles. In this way, Inf-MLLM can preserve the most relevant tokens dynamically and capture long-term dependency during streaming inference (C4). Our contributions are listed as follows. 

\begin{itemize}
   % \item attention saddle, kv cache eviction, 
   % \item attention bias, 
   % \item experiment results

   \item We discover the phenomenon of attention saddles and summarize features of attention patterns on MLLMs. Based on it, we propose an effective KV cache eviction mechanism to reduce memory usage and enable efficient streaming inference of MLLMs on a single GPU.  
   
   % and propose  to involve these tokens in our attention window. Based on them, we design a KV cache eviction mechanism to reduce the computational cost and memory consumption.

   \item We introduce attention bias to update KV cache for long context reasoning. It helps Inf-MLLM to handle streams of texts and videos and capture long-term dependency.

   \item Experiments show that Inf-MLLM facilitates efficient and high-quality streaming inference for multi-round conversations and video clips on edge devices. 
\end{itemize}

% \begin{figure*}[!hbtp]
%     \centering
%     \begin{minipage}{0.21\textwidth}
%         \centering
% 		\includegraphics[width=1.0\textwidth]{figure/heatmaps/head_9_layer_1.pdf}
% 		% \caption{chutian1}
%         {(a) Layer 1}
%     \end{minipage}
%     \begin{minipage}{0.21\textwidth}
%         \centering
% 		\includegraphics[width=1.0\textwidth]{figure/heatmaps/head_9_layer_10.pdf}
% 		% \caption{chutian1}
%         {(b) Layer 10}
%     \end{minipage}
%     \begin{minipage}{0.21\textwidth}
%         \centering
% 		\includegraphics[width=1.0\textwidth]{figure/heatmaps/head_9_layer_19.pdf}
% 		% \caption{chutian1}
%         {(c) Layer 19}
%     \end{minipage}
%      \begin{minipage}{0.21\textwidth}
%         \centering
% 		\includegraphics[width=1.0\textwidth]{figure/heatmaps/head_9_layer_30.pdf}
% 		% \caption{chutian1}
%         {(d) Layer 30}
%     \end{minipage}
%     \caption{Visualization of the attention weight separately in OPT-6.7B (line 1) and Chat-UniVi-7B (line 2) of 30 tokens. The heat maps demonstrate that the attended tokens primarily exhibit the vertical line pattern of relevant tokens and the lower triangular pattern of recent tokens.}
%     \label{heatmaps}
% \end{figure*}

\begin{figure*}[th]
\centering
% \hspace{2mm}
\includegraphics[width=\textwidth]{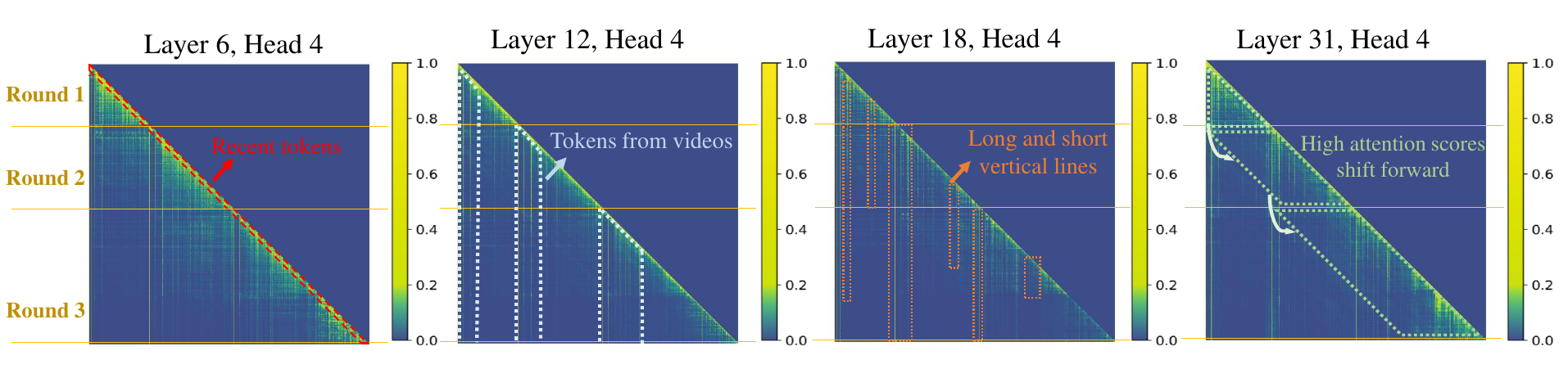}
\caption{Attention maps with typical patterns. We take some layers from the MLLM model, Chat-UniVi-7B, as example. }
\label{patterns}
\end{figure*}

\section{Related Works} \label{sec:related works}

% To enable 

% \paragraph{}

% \paragraph{Context Attention Window.} Context attention window refers to the attended tokens at each token generation step, which is always shorter than the complete context length or the KV cache size. This approach aims to generate the same or similar outputs with fewer tokens while reducing the latency from loading the KV states of attended tokens and the quadratic attention computation. Different designs of context attention window have ranged like window attention, strided window attention, Cswin transformer \cite{dong2022cswin} and Keyformer \cite{adnan2024keyformer}. To get similar outputs, some of these works select tokens following the stationary patterns, while the others dynamically select tokens to constitute the attention window during inference, though they only decrease the number of tokens used in each forward pass and leave all of the tokens store in the memory. Recently, there has been a surge of studies on how to cache tokens according to the context attention window, including StreamingLLM, ${\rm H2O}$ and SnapKV \cite{li2024snapkv}. However, these new methods fail to note the difference between the context attention window with complete KV cache and the reduced KV cache, which falls short of our paper’s primary insight of the attention pattern that ensures discarded tokens will not need to be reused.

\paragraph{KV Cache Eviction} 

Previous works maintain a size-contrained KV cache by evicting KV states of unimportant tokens. Window attention \cite{beltagy2020longformer, jiang2023mistral, liu2022swin, dong2022cswin} caches recent tokens to reduce computation complexity and memory consumption. However, the model performance degrades once the sequence length exceeds the cache size. H2O \cite{zhang2024h2o}, Keyformer \cite{adnan2024keyformer} and SnapKV \cite{li2024snapkv} reduce memory usage with their KV eviction strategy, and H2O enables LLMs to handle texts with infinite length. However, The perplexity is not satisfying on some long text benchmarks due to the improper eviction of important tokens. StreamingLLM \cite{streamingllm} enables LLMs to deal with infinite length by caching the KV states of initial and recent tokens. Although StreamingLLM maintains stable perplexity as the sequence increases in multi-round conversation, it is restricted by its attention window and fails on tasks that demand long-term memory and extensive data dependency, 
such as long document question-answering and long video question-answering. All these methods deal with pure text inputs.

% In this way, they develop different cache management algorithms. As mentioned above, StreamingLLM enables inference on the infinite text by caching only recent tokens and few initial tokens. 
% H2O reduces KV cache memory footprint by up to 10x through a greedy KV cache eviction algorithm. A more recent research, 
% SnapKV \cite{li2024snapkv}, is dedicated to the long input text scenarios and compresses the KV cache of the prompt tokens. 
% Though these works reach good cache reduction results, they don't take effect for our background of streaming inference of MLLMs. These works fall short of our paper’s key insights of the attention patterns on MLLMs in the streaming scenario.

% for computation to overcome the computation and memory latency when modeling long sequences, but 
% still need large memory to store the complete KV cache. 
% incurs model collapse once the sequence length exceeds the cache size.

\paragraph{KV Cache Compression} There exist methods focusing on compressing KV cache. For instance, Transformer-XL \cite{dai2019transformerxlattentivelanguagemodels} splits the entire context into shorter segments with manageable sizes and introduces a recurrence mechanism from RNN to connect adjacent segments. Compressive transformer \cite{rae2019compressivetransformerslongrangesequence} compresses past memories for long-range sequence learning through pooling or convolution. Gear \cite{kang2024gearefficientkvcache} applies dimensionality reduction and quantization to compress the KV cache. 
% \cite{yu2024effectivelycompresskvheads} proves the low-rank property of KV caches and keeps the singular values of KV states instead of the whole KV cache. 
% These works
% performing attention computation and sequence generation with the compressed KV cache or past memory, 
These methods providing another interesting direction to relieve the large memory consumption while achieving good model performance and efficient inference. However, the maximum context length is constrained by the context window determined during pre-training. The compression techniques are orthogonal with KV eviction methods.
% achieve high score on many benchmarks with improved throughput and latency. However, they cannot be applied to deal with streaming LLM inference with long context due to
% pursue computationally similarity but lose the semantic readability of memory. 
% Our work discards tokens gradually to allow for as little information loss as possible, whereas these compression techniques result in unpredictable information loss. 

\paragraph{Relative Position Encoding} Relative position encoding enables LLMs to process longer context during inference while training on shorter texts. Two representative methods are Rotary Position Embeddings (RoPE) \cite{rope} and ALiBi \cite{alibi}.
% which have been applied to numerous LLMs such as Llama2 \cite{llama2}, MPT, Falcon \cite{almazrouei2023falcon} and Pythia \cite{biderman2023pythia}. 
RoPE introduces a rotational encoding method to capture relative token positions.
ALiBi adds negative values 
% that are linear with the inter-token distance 
to weaken the relevance between distant tokens, thus introducing relative position information. 
% to models.
% based on the inter-token distance to the attention score, decreasing larger score for more distant tokens. 
Despite the improvement, their performance declines when the context length exceeds the context window constraint \cite{alibi, chen2023extending}. Although recent works show better performance \cite{peng2023yarn, chen2023extending}, this technique cannot relieve the high memory usage caused by increasing KV states.

\section{Methodology}

\subsection{Attention Patterns of MLLMs}

% Insights

% \subsection{Attention Patterns on LLMs and MLLMs}
We visualize the attention maps of different layers and discover their specific patterns which can benefit the KV cache selection and eviction mechanism. Take the Chat-UniVi-7B \cite{chat_univi} as an example, as shown in Fig. \ref{patterns}. The attention maps of MLLMs exhibit several features. 
% Attention patterns refer to the special patterns exhibited by those attended tokens carrying high attention score. On both LLMs and MLLMs, we discover multiple stable attention patterns following certain rules, which inspires us to design an efficient KV cache mechanism by dynamically involving these tokens.

\noindent\underline{\textit{Pattern 1: recent tokens have high attention scores.}} Recent tokens located at the end of the sequence receive much attention. This is obvious since they are mostly related to the new generated tokens in both position and semantics. 
% These tokens are predictable and easy to reserve.

\noindent \underline{\textit{Pattern 2: tokens converted from videos typically receive hi-}} \underline{\textit{gh attention scores.}}
% Visual modality highlight
We observe an interesting phenomenon that a large number of attention scores are allocated to the region of tokens converted from input videos. For some Vision Language Models (VLMs), the initial tokens of the video even share over 40\% of attention scores. We attribute the feature to the pre-training process, which requires the model to focus on the video content for question answering. 
% Since tokens of the video are normally enclosed by specialized tag tokens in many VLMs, it's easy for the attention score to attend to initial tokens in the video. 
However, since the position of videos is unknown beforehand in the multi-round conversation, an effective method is required to identify important visual tokens dynamically.

\noindent \underline{\textit{Pattern 3: positions with high attention scores appear as ve-}} \underline{\textit{rtical lines.}} 
% Vertical-line pattern}}: 
Besides recent tokens and key visual tokens, we find that high attention scores are also distributed among tokens scattered in the sequence. These tokens 
% carrying high attention scores, 
are attended to for dozens or hundreds of decoding steps, resulting in short or long vertical lines on the attention map. 
% A characteristic of \textbf{continuity} is evident, 
% The vertical lines imply that we can predict the importance of a token based on its attention score at previous decoding steps and decide whether to cache the token.
A special case is the attention sink named by StreamingLLM \cite{streamingllm}, which refers to the initial tokens because they are endowed with huge attention score by SoftMax. Unlike StreamingLLM that only caches static initial tokens, Inf-MLLM can dynamically identify the influential scattered tokens, including the initial tokens.

\noindent \underline{\textit{Pattern 4: high attention scores shift forward as the multi-}} \underline{\textit{round inference progresses.}}
% Transfering attention focus in streaming inferen-}} \underline{\textit{ce}}: 
During streaming inference, we observe that high attention scores shift forward across conversation rounds. When a new prompt comes, the distribution of attention scores changes significantly, indicating that the attention window containing attended tokens should be updated correspondingly, especially when a new conversation round starts. 
% When a new prompt is input to the model, the attention patterns largely change and the attention score always moves afterwards. 
% It exposes that in streaming inference, an attention window containing attended tokens should be updated frequently, especially when each new conversation round starts. 
Existing methods cannot capture the shifting feature and simply accumulate attention scores for KV selection, making large scores aggregate at earlier tokens while ignoring important newer tokens.
% by their accumulated attention score and introduce serious imbalance, since accumulated attention score favors prior tokens by default.

\begin{figure}[ht]
\centering
% \hspace{2mm}
\includegraphics[width=0.99\columnwidth]{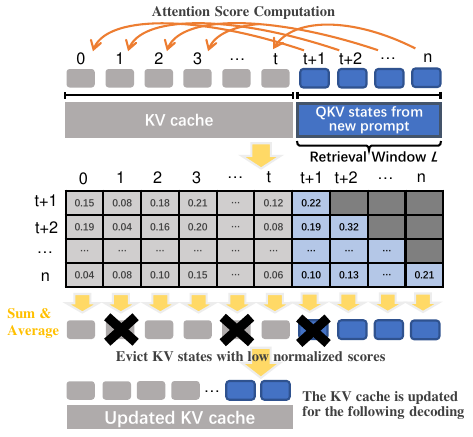}
\caption{The illustration of KV cache eviction. It happens when a new prompt comes during streaming inference.}
\label{fig:cacheevict}
\end{figure}

\begin{figure*}[t]
\centering
% \hspace{-5mm}
\includegraphics[width=0.99\textwidth]{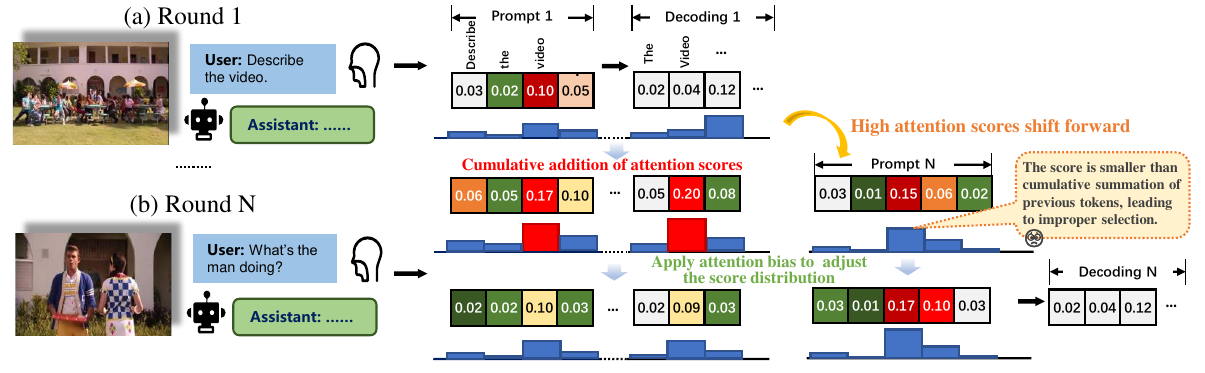}
\caption{The illustration of attention bias to adjust the distribution of attention scores during streaming inference.}
\label{bias}
\end{figure*}

The attention patterns present tokens with high attention scores which are most relevant to the decoding of the current token. We term these tokens as \textbf{attention saddles}, borrowing the concept of saddle points in mathematics. To identity and always maintain attention saddles in KV cache, we proposes two techniques as follows. 
% To utilize these patterns, we propose attention saddles which cover the tokens in Pattern 2 to Pattern 4, and formulate the attention window consisting of recent tokens and attention saddles. We identify attention saddles through following approaches:

\begin{itemize}
    \item We design a KV cache eviction mechanism to evict irrelevant KV states while maintaining attention saddles (pattern 1, 2 and 3) in KV cache. 
    
    % retrieval window consisting of recent tokens at the end of the sequence to evaluate average attention score of remaining tokens (Pattern 2). Since the rear tokens are most close to new generated tokens, they can predict the attended tokens in following decoding steps utilizing the continuity (Pattern 3).

    \item We introduce an attention bias to dynamically update the KV cache and capture the shifting feature (pattern 4), enabling MLLMs with long-term memory.
    % evicting KV states of irrelevant earlier tokens while preserving new attention saddles. 
    % By keeping most relevant tokens in the cache during streaming inference, Inf-MLLM can enable MLLMs with long-term memory.  
    % To offset the inherent imbalance in attention score, we introduce attention bias to update the KV cache (attention window) and force the models to always focus on the newest context (Pattern 4). Through regulating the attention bias, we can further enable models to capture the long-term dependency.
\end{itemize}

% \begin{figure*}[tp]
% \centering
% \hspace{-5mm}
% \includegraphics[width=0.90\textwidth]{figure/f3.pdf}
% \caption{The streaming attention pattern of language models. As the new prompt arrives and a new iteration of inference begins, the focus of attention score gradually moves backwards, implying that prior tokens generated in the rounds long ago can be discarded from the cache to save memory and accelerate decoding speed.}
% \label{conversation}
% \end{figure*}

\subsection{KV cache eviction and updating}
% Dynamically Predict Attended Tokens Based on Continuity
Inf-MLLM employs an efficient KV cache eviction and updating mechanism, as shown in Fig. \ref{fig:cacheevict}. During the streaming inference, when a new prompt comes, the QKV states of each token in the prompt are computed and stored in a retrieval window with a length of $L$. Suppose there exist KV states of $t$ earlier tokens from previous rounds of inference. Attention scores are computed by multiplying queries of the $L$ new tokens with the KV states of $t$ tokens in the cache and the KV states of the new tokens themselves, generating a $L*n$ matrix as shown in Fig. \ref{fig:cacheevict}. 

To identify attention saddles and evict KV states of irrelevant tokens, the attentions scores for each token are accumulated. Due to the continuity of vertical lines exhibited in attention patterns, we conduct a local sum within the retrieval window length (from the (t+1)-th to the n-th row) to improve computation efficiency, rather than aggregate along the complete attention matrix. The summation results are then normalized to avoid the accumulation of attention scores in earlier tokens. This is adaptive to the shifting feature of the attention pattern.  
% enable the infinite context modeling. To identify KV states, we adopt an intuitive but effective policy that to evaluate KV states with the tokens at the end of the sequence to utilize the continuity. 
% In this paper, we update the KV cache at prefilling phases in the streaming inference, and evaluate with $l$ tokens from the new prompt, referred to as the retrieval window. 
% As in Figure \ref{window}, the algorithm calculates the attention weight between the QKV states of the retrieval window and the KV states from remaining tokens outside the window and the KV cache. 
% To reduce the additional cost of our algorithm, we reuse the computation result of the model in the forward pass. 
% After getting the attention weight, we sum the weight and takes the average along the direction of the retrieval window. 
After that, Inf-MLLM select the KV states of tokens with top-$t$ highest normalized attention scores and evict less important KV states from the cache.
% Based on the average weight, we select top-$r$ tokens (KV states) with highest attention scores and evict others. 
The updated KV cache will be utilized for the following decoding process at the current round. Inf-MLLM invokes the KV cache eviction and updating mechanism at the beginning of each conversation round when a new prompt arrives, and does not evict tokens during decoding steps. Therefore, the cost of KV cache eviction and updating is negligible, and the inference speed of models is increased since fewer tokens are involved in computations after eviction.

\subsection{Attention Bias} 
% : Update KV Cache and Regulate Long-term Dependency

% In the last section, we propose our KV cache updating algorithm that evicts KV states with the retrieval widnow and new prompt. 
To further strengthen the ability of KV cache eviction, especially in long context processing and multi-round conversations, some issues need to be solved.  
% Though avoiding the interference of outdated information, we still have to address the imbalance inherently lying in attention score. The imbalance is raised from two main aspects. 
Firstly, because of the SoftMax operation, the total sum of attention scores or weights maintains as one, despite the increasing sequence length and the growing number of tokens. This means that the weight of each token degrades gradually as the inference progresses, and the difficulty of identification for high-score tokens is exacerbated. Secondly, 
% there exist imbalance in attention s is even more serious when employing a size-constrained KV cache with eviction mechanism. 
after several rounds of KV eviction, the distribution of attention scores becomes uneven among the remaining tokens and the attention score of some tokens can be enhanced due to multiple rounds of accumulation, as shown in Fig. \ref{bias}. This phenomenon can prevent the identification of new attention saddles which are more relevant to the current conversation round, leading to improper KV eviction and even model collapse when the cache is almost not updated after rounds of inference on long context.
% may enhance the score of cached prior tokens, and probably forms a vicious cycle that cached tokens squeeze out new tokens and scale up their score further. This finally causes the models to collapse when the cache is almost no longer updated on the long context. 

To update the KV cache continuously in streaming inference, we introduce attention bias to shift the attention focus to the newest context. We demonstrate its effects in Fig. \ref{bias}, where attention bias can adjust the distribution of attention scores and enables the multi-round video conversation to continue. The attention bias is employed when identifying the attention saddles. After calculating the average attention scores in retrieval window, we add the attention bias to them to impel the KV cache to discard tokens retained long ago. With the higher attention bias, the KV cache tends to involve more new tokens and the model focuses more on the incoming tokens to adapt to streaming scenarios. With relatively lower attention bias, the KV cache can retain prior tokens longer and the model is able to capture longer-term dependency. Therefore, properly adjusting the attention bias can preserve long-term dependency while ensuring long context streaming inference. 

\subsection{Inf-MLLM Algorithm}

In this section, we present the overview of our Inf-MLLM algorithm. We highlight its core idea in maintaining a size-restrained KV cache consisting of recent tokens and relevant tokens based on the attention patterns we have observed and implement our KV cache eviction mechanism with the retrieval window and attention bias. We also employ the length extrapolation techniques to deal with the overlong context exceeding pre-training length of the model. Inf-MLLM is able to be applied on streaming scenarios where text and video inputs are streamed in and need to processed continuously. The details are provided in Algorithm \ref{alg:algorithm1}.

\begin{algorithm}[t]
\caption{The overview of the Inf-MLLM algorithm}
\label{alg:algorithm1}
\textbf{Input}: Attention score matrix $W \in \mathbf{R}^{m\times n}$, KV states $ K,V \in \mathbf{R}^{n\times d}$, retrieval window size $l$, number of relevant tokens $r$, attention bias $b$. Note that though the algorithm below is applied on the KV cache of one layer, Inf-MLLM in fact process KV cache of all layers simultaneously utilizing the parallelism of PyTorch.    \\
\textbf{Output}: KV cache ($K_s$, $V_s$)
\begin{algorithmic}[1] %[1] enables line numbers
\STATE $S$ = $\frac{1}{l}\Sigma W[m-l : m, 0 : n-l] $ \COMMENT{$S \in \mathbf{R}^{n-l}$}
\STATE $d$ = $b/(n-l)$  \COMMENT{Attention bias parameter}
\STATE $D$ = $-[n-l-1, \cdots, 0] * d$ \COMMENT{Attention bias}
\STATE $W$ = $S + D$   \COMMENT{Biased attention score}
\STATE $I_r$ = $Top_k(W, r)$ \COMMENT{Indices of relevant tokens}
\STATE $I_l$ = $[n-l, \cdots, n]$ \COMMENT{Indices of recent tokens}
\STATE $I$ = $[I_r, I_l]$ \COMMENT{$I \in \mathbf{R}^{r+l}$}
\STATE $K_s, V_s = K[I, :], V[I, :]$ \COMMENT{Compress KV cache}
\STATE \textbf{return} ($K_s$, $V_s$)
\end{algorithmic}
\end{algorithm}

\begin{figure*}[ht]
\centering
    \begin{minipage}{0.33\textwidth}
        \centering
		\includegraphics[width=1.0\textwidth]{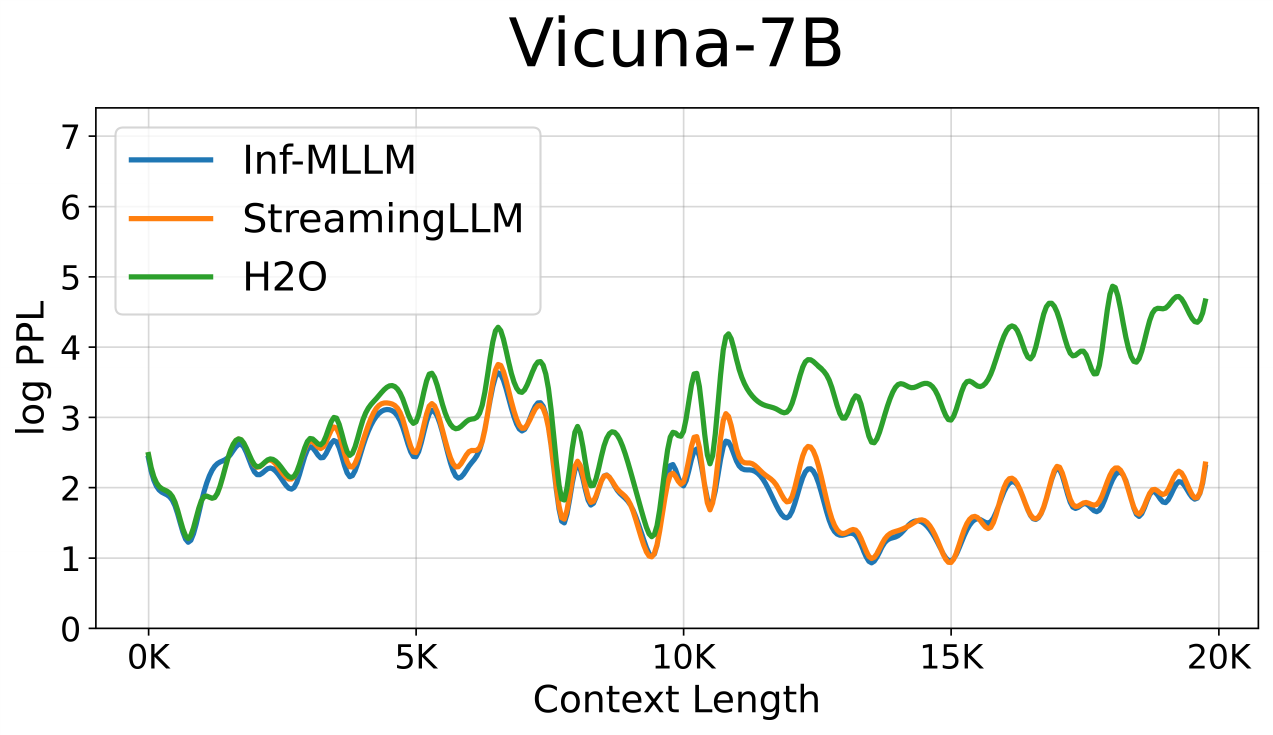}
		% \caption{chutian1}
    \end{minipage}
    \begin{minipage}{0.33\textwidth}
        \centering
		\includegraphics[width=1.0\textwidth]{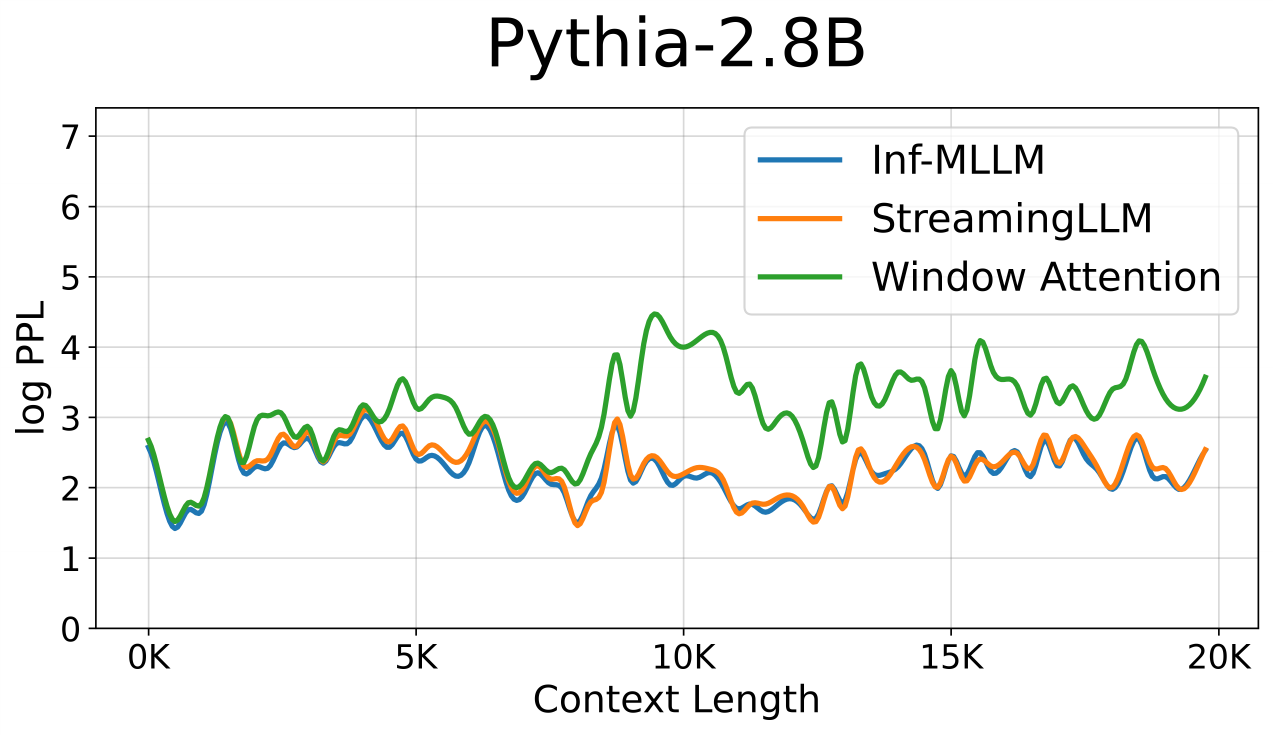}
		% \caption{chutian1}
    \end{minipage}
    \begin{minipage}{0.33\textwidth}
        \centering
		\includegraphics[width=1.0\textwidth]{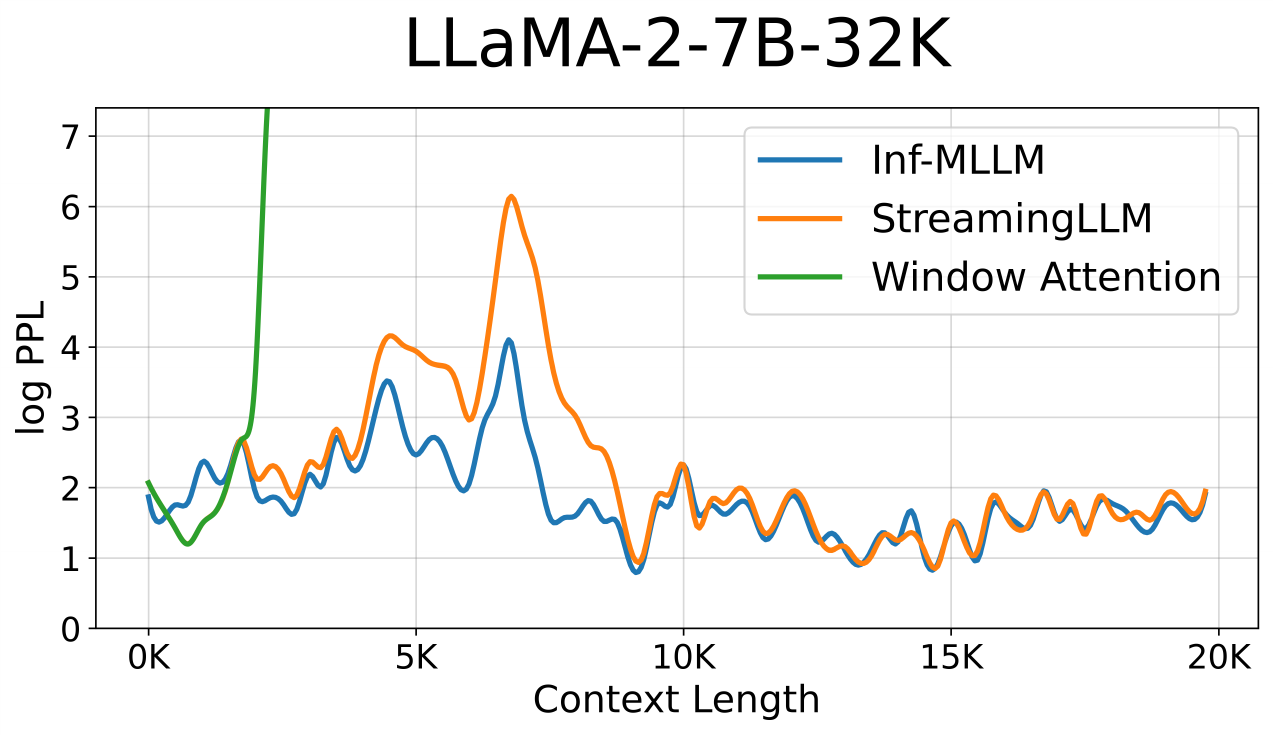}
		% \caption{chutian1}
    \end{minipage}
  %    \begin{minipage}{0.32\textwidth}
  %       \centering
		% \includegraphics[width=1.0\textwidth]{figure/bar/Vicuna-7B.pdf}
		% % \caption{chutian1}
  %   \end{minipage}
  %   \begin{minipage}{0.32\textwidth}
  %       \centering
		% \includegraphics[width=1.0\textwidth]{figure/bar/Pythia-2.8B.pdf}
		% % \caption{chutian1}
  %   \end{minipage}
  %   \begin{minipage}{0.32\textwidth}
  %       \centering
		% \includegraphics[width=1.0\textwidth]{figure/bar/LLaMA-2-7B-32K.pdf}
		% % \caption{chutian1}
  %   \end{minipage}
     \begin{minipage}{0.33\textwidth}
        \centering
		\includegraphics[width=1.0\textwidth]{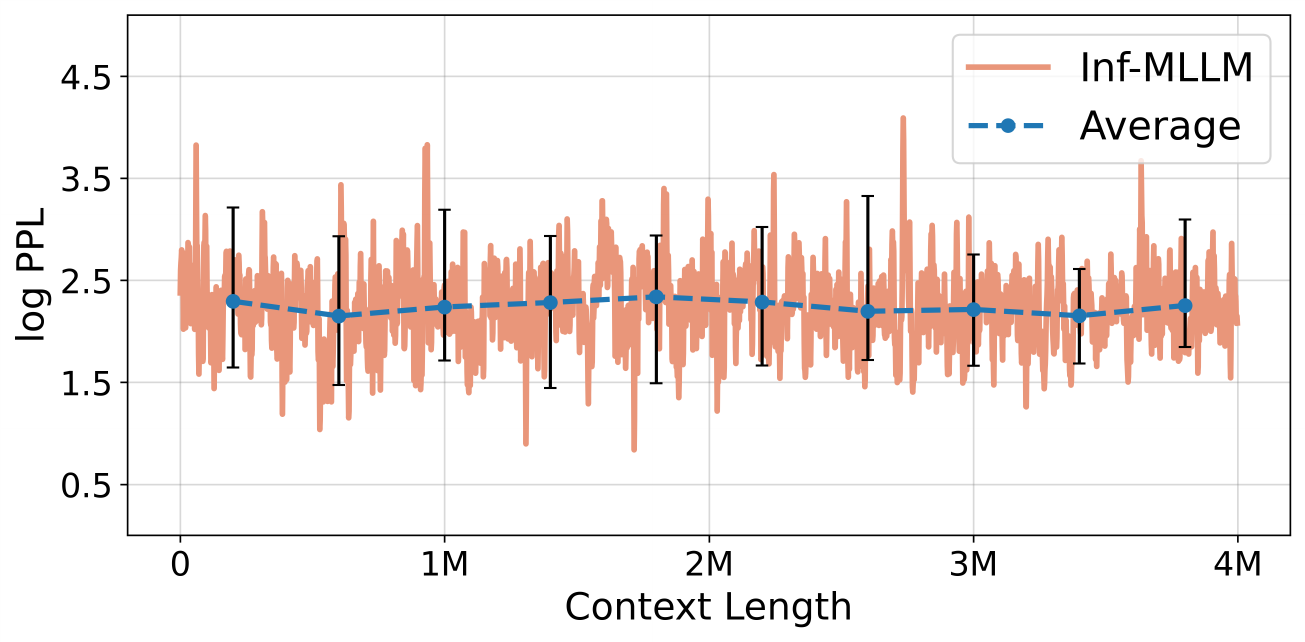}
		% \caption{chutian1}
    \end{minipage}
    \begin{minipage}{0.33\textwidth}
        \centering
		\includegraphics[width=1.0\textwidth]{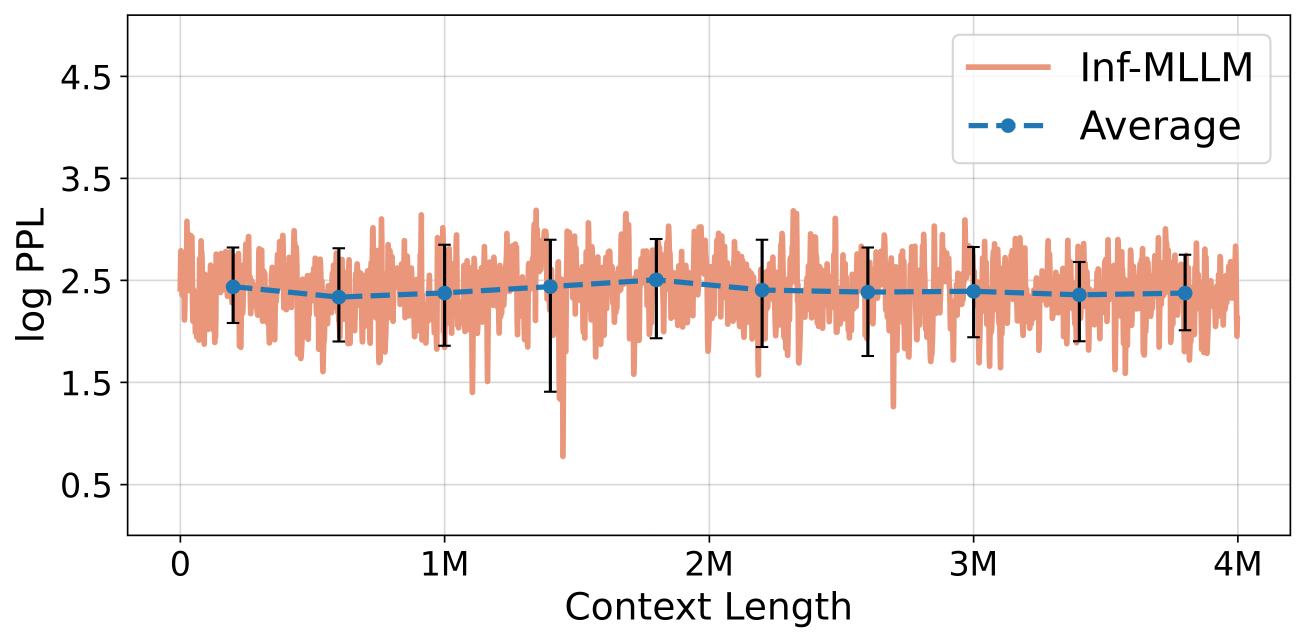}
		% \caption{chutian1}
    \end{minipage}
    \begin{minipage}{0.33\textwidth}
        \centering
		\includegraphics[width=1.0\textwidth]{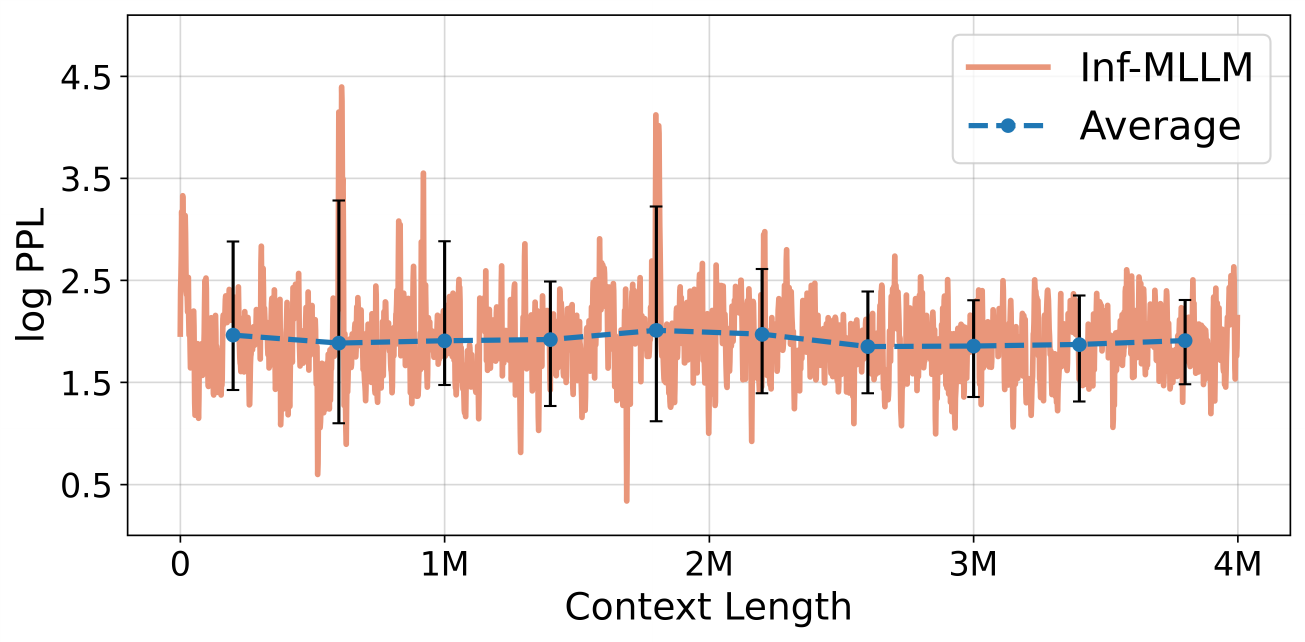}
		% \caption{chutian1}
    \end{minipage}
\caption{LLM perplexity comparison on the Wiki-Text-103 dataset with different context lengths.}
% for Vicuna-7B, Pythia-2.8B and LLaMA-2-7B-32K, given different content lengths. 
% Line 1: comparison results between various techniques including window attention, H2O, StreamingLLM and our Inf-MLLM on texts with 20K tokens. 
% Line 2: performance of different techniques with varying KV cache size. 
% Line 3: perplexity of Inf-MLLM for long texts with 4 million tokens across various LLMs.}
\label{eval_text}
\end{figure*}

% \subsection{Long Prompt Segmentation Mechanism}

% \begin{figure}[H]
% \centering
% \includegraphics[width=0.95\columnwidth]{figure/nll.png} 
% \caption{Language modeling perplexity of Inf-MLLM on super long texts with 1 million tokens across Vicuna and Pythia.}
% \label{nll}
% \end{figure}

\section{Experiments}

% In this section, our goal is to prove that Inf-MLLM is able to support language modeling on overlong context of different modalities for MLLMs. We demonstrate that Inf-MLLM, with many innovative designs, performs better than previous methods.

% \begin{itemize}
%     \item In Section 4.1, we evaluate 
%     \item In Section 4.2,
%     \item In Section 4.3,
% \end{itemize}

We evaluate Inf-MLLM on both LLMs and MLLMs with pure texts and texts/videos as input. 
% to show its ability to support multimodal streaming inference for MLLMs. 
We test on three prominent LLMs, namely Vicuna-7B \cite{vicuna}, Pythia-2.8B \cite{biderman2023pythia} and LLaMA-2-7B-32K \cite{LLaMA-2-7B-32K}, and two state-of-the-art MLLMs for videos, namely Chat-UniVi-7B \cite{chat_univi} and Flash-VStream-7B \cite{zhang2024flashvstreammemorybasedrealtimeunderstanding}. All of these models are employed with relative position encoding such as RoPE \cite{rope}. For pure text inputs, we compare Inf-MLLM with typical baselines including window attention \cite{beltagy2020longformer}, H2O \cite{zhang2024h2o} and StreamingLLM \cite{streamingllm}. For video and text inputs, we evaluate the streaming inference performance of MLLMs empowered with and without Inf-MLLM. 
% demonstrate that Inf-MLLM enables MLLMs to provide stable performance without accuracy drop while raw models fail due to Out-of-Memory (OOM) error. 
All experiments are conducted on a single NVIDIA 4090D GPU or NVIDIA ORIN GPU, 
% with models of appropriate size, 
demonstrating the powerful capability of Inf-MLLM on resource-constrained devices.

\subsection{LLM Perplexity on Super Long Texts}

We first compare Inf-MLLM with previous methods in LLM perplexity on long text inputs, as shown in Fig. \ref{eval_text}.
% We evaluate Inf-MLLM's long text modeling perplexity using the Wiki-Text-103 dataset along with other techniques across three LLMs: Vicuna-7B, Pythia-2.8B and LLaMA-2-7B-32K. 
The maximum context lengths of the tested LLMs, Vicuna-7B, Pythia-2.8B, and LLaMA-2-7B-32K, are 2K, 2K and 32K, respectively. After applying KV cache eviction strategies, the context length can be extended.  
% which requires length extrapolation techniques to model longer texts. Though LLaMA-2-7B-32K is able to model up to 32K tokens, edge devices are hard to afford the huge memory budget without Inf-MLLM. In all subsequent experiments with Inf-MLLM, the KV cache size is set by default 2K.
% Figure \ref{eval_text} illustrates our detailed experimental results. 
We can see that for context length up to 20K, Inf-MLLM reaches better perplexity than window attention, H2O and StreamingLLM. Note that H2O only supports Vicuna-7B and the perplexity of window attention increases rapidly when exceeding the 2K limit on LLaMA-2-7B-32K.
% By the way, the raw models result in Out-of-Memory (OOM) error on such long texts. 
% Varying the KV cache size from 500 to 2000, Inf-MLLM remains relatively low perplexity which is always lower than baselines. 
We also evaluate Inf-MLLM on texts with up to 4 million tokens, as shown in Fig. \ref{eval_text}.
% To further test its performance on super long texts, we evaluate Inf-MLLM on texts with over 4 million tokens. 
The results show that the LLMs empowered with Inf-MLLM presents stable perplexity on super long text inputs, which largely surpass the maximum context length constraint.
% can reliably handle exceptionally extended texts encompassing more than 4M tokens with stable perplexity, across a spectrum of LLMs on a single GPU. 

\subsection{Long-term Memory Capability} \label{sec:long-term mem}

To evaluate the capability of long-term memory, we design a multi-round question-answering benchmark based on the LongEval-LineRetrieval dataset \cite{longchat2023}. The dataset involves 300 prompts each of which contains multiple lines of texts in the format of ``The REGISTER\_CONTENT in line $index$ is $number$", and requires models to answer the $number$ given $index$ at the end of the prompt. We vary the distance between the final question and the corresponding answer line to evaluate the ability of long-term memory. 
% This benchmark is especially persuasive for techniques that continuously discard tokens from KV cache and reshape the attention weight, as we analyze in Section 3.3.

% \paragraph{Setup.} Our experiments are based on three different models: Vicuna-7B \cite{vicuna}, Pythia-7B \cite{biderman2023pythia} and LLaMA-2-7B-32K \cite{LLaMA-2-7B-32K}. We evaluate models on Wiki-Text-103 and LongEval-LineRetrieval dataset to test the performance of Inf-MLLM on long texts. We use NVIDIA 4090D GPU.

We select StreamingLLM (abbreviated to StrLLM) as the baseline since it outperforms other previous methods on long text inputs. As shown in Table \ref{eval_long}, 
% compare Inf-MLLM with StreamingLLM (abbreviated to StrLLM) on Vicuna-7B and LLaMA-2-7B-32K, since StreamingLLM performs best in the long text evaluation.
Inf-MLLM reaches higher accuracy across all token distances and LLMs. The superiority is particularly significant on LLaMA-2-7B-32K, where we set the 
% proportion of attention saddles in the cache to 80\% and 
attention bias to 0.0001. Inf-MLLM maintains close to 100\% accuracy while StreamingLLM drops to less than 50\% at different token distances. The improvement can be attributed to 
% The reason behind the big difference includes two main points that 
(i) the relevant tokens broaden the span of attention window and (ii) the attention bias compensates the reshaped attention scores. Therefore, Inf-MLLM presents stable streaming performance with longer-term memory compared to existing methods.

\begin{table*}[ht]
\centering
\caption{Comparison on zero-shot multi-round video question-answering tasks (5, 10 and 300 rounds on each task). \textcolor{blue}{OOM}: Out-of-Memory. 
% as the number of rounds increases. 
The KV cache size is 2K. Evaluation is based on protocols (accuracy and score)
% We follow the protocol to evaluate the results 
using GPT-3.5-Turbo-1025.} 
% and list the accuracy and score in the table. }
% \renewcommand{\arraystretch}{1.4}
\setlength{\tabcolsep}{1mm}
  \small
\begin{tabular}{l|cccccc|cccccc|cccccc}
\Xhline{1.2pt}
\multicolumn{1}{c|}{}                                & \multicolumn{6}{c|}{\textbf{MSVD-QA}}                                         & \multicolumn{6}{c|}{\textbf{MSRVTT-QA}}                                       & \multicolumn{6}{c}{\textbf{TGIF-QA}}                                     \\
\multicolumn{1}{c|}{}                                & \multicolumn{3}{c}{{\textbf{Accuracy}}} & \multicolumn{3}{c|}{{\textbf{Score}}}     & \multicolumn{3}{c}{{ \textbf{Accuracy}}} & \multicolumn{3}{c|}{{\textbf{Score}}}     & \multicolumn{3}{c}{{\textbf{Accuracy}}} & \multicolumn{3}{c}{{\textbf{Score}}} \\ 
\cmidrule(lr{2pt}){2-4} \cmidrule(lr{2pt}){5-7} \cmidrule(lr{2pt}){8-10} \cmidrule(lr{2pt}){11-13} \cmidrule(lr{2pt}){14-16} \cmidrule(lr{2pt}){17-19}
\multicolumn{1}{c|}{\multirow{-3}{*}{\textbf{Models}}} & 5            & 10         & 300        & 5         & 10         & 300         & 5           & 10          & 300       & 5         & 10         & 300         & 5           & 10          & 300       & 5        & 10       & 300       \\ 
\Xhline{1.2pt}
Chat-UniVi (w/o ours)                                        & 60.0         & 70.0       & \textcolor{blue}{OOM}         & 3.8       & 4.0        & \textcolor{blue}{OOM}         & 20.0        & 40.0        & \textcolor{blue}{OOM}         & \textbf{2.8}       & \textbf{3.1}        & \textcolor{blue}{OOM}          & \textbf{80.0}        & 70.0        & \textcolor{blue}{OOM}         & \textbf{4.4}      & \textbf{4.0}      & \textcolor{blue}{OOM}        \\
Chat-UniVi (w/ ours)                               & \textbf{100.0}        & \textbf{90.0}       & \textbf{72.7}       & \textbf{4.4}       & \textbf{4.1}        & \textbf{3.9}         & \textbf{40.0}        & \textbf{40.0}        & \textbf{53.3}       & 2.6       & 2.8        & \textbf{3.34}        & 60.0        & \textbf{70.0}        & \textbf{67.0}       & 3.6      & 3.8      & \textbf{3.90}      \\
\hline
Flash-VStream (w/o ours)                                    & 80.0         &  \multicolumn{2}{c}{\textcolor{blue}{OOM}} 
& 3.8       & \multicolumn{2}{c|}{\textcolor{blue}{OOM}} 
& 20.0        & \multicolumn{2}{c}{\textcolor{blue}{OOM}}  
& \textbf{2.8}       & 
% \textcolor{blue}{OOM} & \textcolor{blue}{OOM} 
\multicolumn{2}{c|}{\textcolor{blue}{OOM}} 
& 60.0        & \textbf{50.0}        & \textcolor{blue}{OOM}         & \textbf{3.2}      & 3.0      & \textcolor{blue}{OOM}        \\
Flash-VStream (w/ ours)                           & \textbf{100.0}        & \textbf{90.0}       & \textbf{65.7}       & \textbf{5.0}       & \textbf{4.4}        & \textbf{3.54}        & \textbf{20.0}        & \textbf{40.0}        & \textbf{54.0}       & 1.6       & \textbf{2.5}        & \textbf{3.16}        & \textbf{60.0}        & 40.0        & \textbf{63.7}       & 3.0      & \textbf{3.1}      & \textbf{3.7}       \\ 
\Xhline{1.2pt}
\end{tabular}
\label{eval_video}
\end{table*}

% Please add the following required packages to your document preamble:
% \usepackage{multirow}
\begin{table}[ht]
\caption{Accuracy comparison on the LongEval-LineRetrieval dataset. 
% With increasing token distances, we compare the performance of Inf-MLLM and StreamingLLM on Vicuna-7B and LLaMA-2-7B-32K. 
% Each line contains 23 tokens.
Higher values mean better accuracy.}
% StrLLM denotes StreamingLLM.}
\centering
\label{eval_long}
\setlength{\tabcolsep}{1mm}
  \small
\begin{tabular}{cc|cccc}
\Xhline{1.2pt}
\multirow{2}{*}{\makecell[c]{\textbf{Line} \\ \textbf{Distance}}} & \multirow{2}{*}{\makecell[c]{\textbf{Token} \\ \textbf{Distance}}} & \multicolumn{2}{c}{\textbf{Vicuna-7B}} & \multicolumn{2}{c}{\textbf{LLaMA-2-7B-32K}} \\ 
\cmidrule(r){3-4} \cmidrule(l{3pt}){5-6}
% \cline{3-4} \cline{5-6}
                &           & \textbf{StrLLM}  & \textbf{Ours}  & \makebox[0.14\columnwidth][c]{\textbf{StrLLM}}     & \makebox[0.14\columnwidth][c]{\textbf{Ours}}      \\
\Xhline{1.2pt}
5               & 115               & 0.98        & \textbf{0.98}     & 0.40      & \textbf{1.00}        \\
10              & 230               & 0.97        & \textbf{0.98}     & 0.07      & \textbf{0.99}           \\
15              & 345               & 0.90        & \textbf{0.98}     & 0.04      & \textbf{0.99}       \\
20              & 460              & 0.80         & \textbf{0.92}     & 0.51      & \textbf{1.00}        \\
25              & 575              & 0.76         & \textbf{0.88}     & 0.22      & \textbf{0.99}       \\
30              & 690              & 0.79         & \textbf{0.90}     & 0.07      & \textbf{0.87}       \\
35              & 805             & 0.70          & \textbf{0.73}     & 0.02      & \textbf{0.99}     \\
\Xhline{1.2pt}
\end{tabular}
\end{table}

\subsection{Multi-round Video Question-answering}

Inf-MLLM enables efficient streaming inference for MLLMs on overlong multimodal inputs such as videos. We test Inf-MLLM on two state-of-the-art Vision Language Models (VLMs), Chat-UniVi and Flash-VStream, using three popular video question-answering datasets including MSVD-QA \cite{msvd}, MSRVTT-QA \cite{msrvtt} and TGIF-QA \cite{jang2017tgifqaspatiotemporalreasoningvisual}. We formulate three multi-round video question-answering benchmarks by concatenating each sample in three datasets. 

As shown in Table \ref{eval_video}, Inf-MLLM improves the model performance for most cases and extensively enables models to continuously process new video clips and maintain high-quality answering up to 300 rounds of conversations. The original models fail at long contexts due to out-of-memory (OOM). 
% On the MSVD-QA dataset, models with Inf-MLLM always surpass baseline models when the number of rounds is small, achieving 100\% for 5 rounds and 90\% for 10 rounds in average accuracy. As the video conversation proceeds, baselines fail for the OOM error, while Inf-MLLM enables the two models to continue processing new video clips and questions with reasonable accuracy and quality (score). 
Although these VLMs compress and truncate patch tokens based on similarity between video frames, the memory usage issue will still be severe due to the increasing KV states in the streaming inference. Inf-MLLM successfully solves this issue due to its effective KV eviction mechanism which maintains a small size of KV cache (2K). Note that despite the slight decrease of the score metric in some cases, models with Inf-MLLM can still provide correct answers while incurring minor issues like description redundancy.

% Recently, researches on compressing videos utilizing the similarity is prevalent in the field of VLM. These works, such as Chat-UniVi and Flash-VStream, mainly use clustering algorithm to truncate similar ViT patch tokens from the video. However, they are still hard to tackle long videos or long video question-answering scenarios as our experiment shows. As a result, our Inf-MLLM is instructive for future VLMs. 

% Please add the following required packages to your document preamble:
% \usepackage{multirow}
% \usepackage[table,xcdraw]{xcolor}
% Beamer presentation requires \usepackage{colortbl} instead of \usepackage[table,xcdraw]{xcolor}
\begin{table}[h]
\centering
\caption{Evaluation on the VStream-QA benchmark. The video length means the conversation is around the video clips of that time slot in the video. 
% The results prove that Inf-MLLM maintains reasonable accuracy and score on long videos. 
}
\setlength{\tabcolsep}{1mm}
\small
\begin{tabular}{l|cccccc}
\Xhline{1.2pt}
\multicolumn{1}{c|}{}                           & \multicolumn{6}{c}{\textbf{VStream-QA}}                                        \\
\multicolumn{1}{c|}{\multirow{-2}{*}{}}         & \multicolumn{3}{c}{\textbf{Accuracy}} & \multicolumn{3}{c}{\textbf{Score}}  \\
\cmidrule(r){1-1} \cmidrule(lr{2pt}){2-4} \cmidrule(lr{2pt}){5-7}
% \cline{1-1} \cline{2-4} \cline{5-7}
\multicolumn{1}{l|}{\textbf{Round}}      & 2          & 4          & 300         & 2         & 4          & 300        \\
\hline
\multicolumn{1}{l|}{\textbf{Video Length (min)}} & 2.83       & 3.22       & 67.35       & 2.83      & 3.22       & 67.35      \\ \Xhline{1.2pt}
Chat-UniVi (w/o ours)                                    & 50.0       & \multicolumn{2}{c}{\textcolor{blue}{OOM}}  & 3.5       & \multicolumn{2}{c}{\textcolor{blue}{OOM}} \\
Chat-UniVi (w/ ours)                                 & \textbf{50.0}       & \textbf{25.0}       & \textbf{37.7}        & \textbf{3.5}       & \textbf{3.3}        & \textbf{3.0}        \\ \hline
Flash-VStream (w/o ours)                                 & 50.0       & 50.0       & \textcolor{blue}{OOM}         & 3.0       & 2.5        & \textcolor{blue}{OOM}        \\
Flash-VStream (w/ ours)                             & \textbf{50.0}       & \textbf{50.0}       & \textbf{40.7}        & \textbf{3.5}       & \textbf{3.5}        & \textbf{3.2}        \\ \Xhline{1.2pt}
\end{tabular}
\label{eval_stream}
\end{table}

\begin{figure*}[t]
\centering
% \hspace{-5mm}
\includegraphics[width=\textwidth]{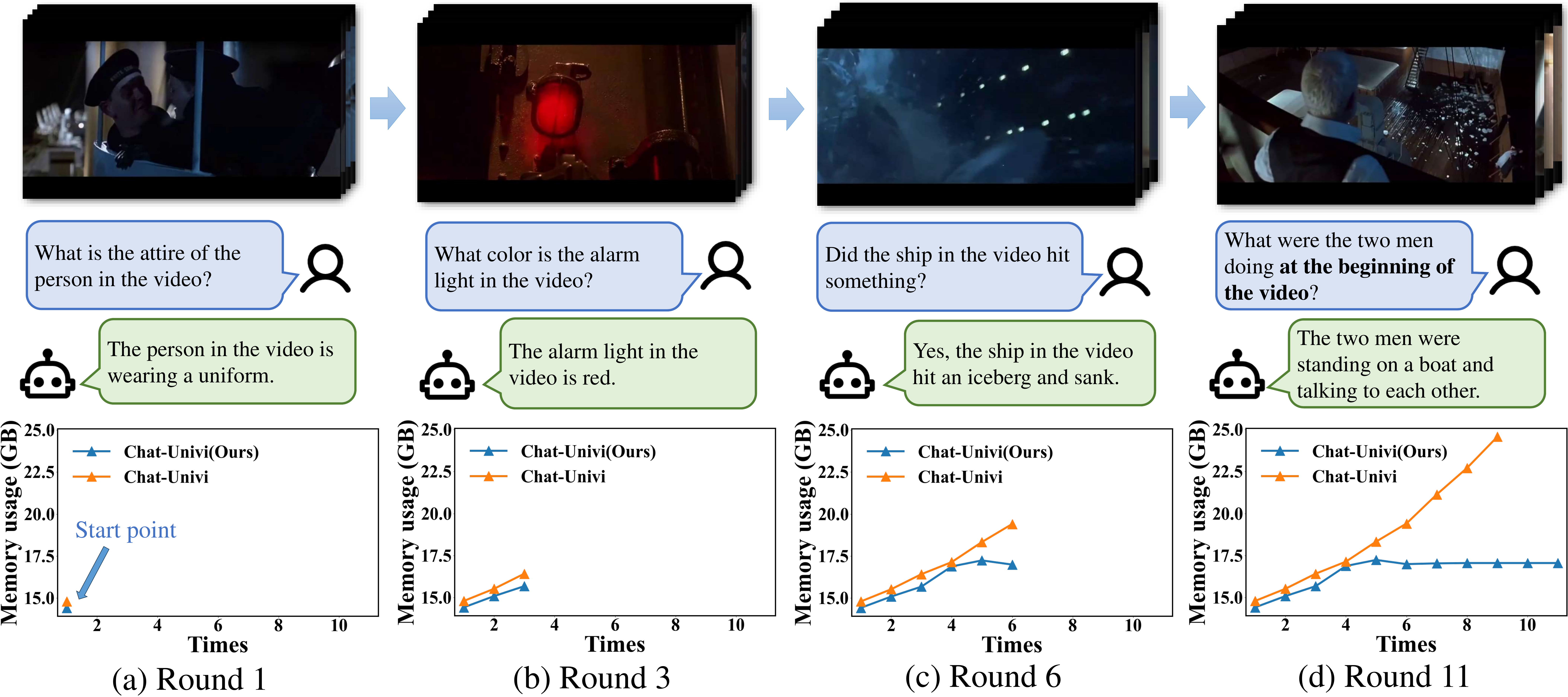}
\caption{Inf-MLLM equips large video models with the ability to manage long videos and engage in multi-turn conversations. We deploy a multimodal chatbot on Orin, which asks the chatbot a question every 30 seconds while playing a video. This example excerpts the dialogue from rounds 1, 3, 6, and 11. The bottom graph illustrates the memory usage comparison between the multimodal chatbot deployed with Inf-MLLM and the original version of Chat-UniVi. In this example, we are playing a clip from the movie Titanic, which lasts for six and a half minutes.}
\label{demo}
% \vspace{-0.3cm}
\end{figure*}

\subsection{Question-answering for Long Video Streams}

We also test Inf-MLLM on a recently released benchmark, VStream-QA \cite{zhang2024flashvstreammemorybasedrealtimeunderstanding}, which focuses on online video stream understanding. VStream-QA includes extremely long videos that last from 30 minutes to over 1 hour. Each sample contains video clips of around 20 seconds to 5 minutes. Similarly, we test Inf-MLLM using Chat-UniVi-7B and Flash-VStream-7B. Table \ref{eval_stream} shows that Inf-MLLM enables models to deal with long video streams and continuously generate high-quality answers, even as the video length grows to over 1 hour and the length of context comprised of both video clips and texts increases to up to 220K.

\subsection{Efficiency Evaluation}

\begin{figure}[t]
% \centering
% \hspace{-1mm}
\includegraphics[width=0.99\columnwidth]{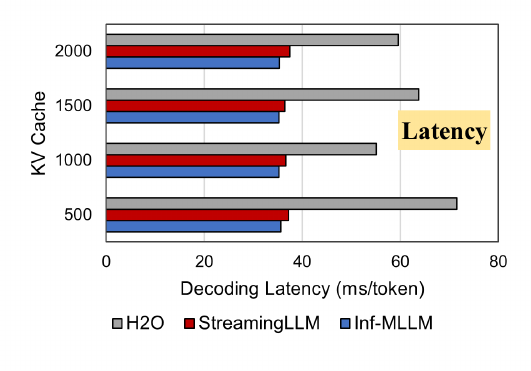} 
\caption{Comparison of the decoding latency when varying the KV cache size on the Y-axis. 
% Inf-MLLM reaches better performance on all cache sizes than H2O and StreamingLLM baselines.
}
\label{memory_latency}
\end{figure}

We evaluate the efficiency of different methods in terms of the decoding latency and memory usage on a NVIDIA 4090D GPU using the Vicuna-7B model, as shown in Fig. \ref{memory_latency}.
% We especially focus on the benchmark against H2O since it employs complex cache management mechanism like Inf-MLLM. 
Compared to other methods, Inf-MLLM achieves stably smaller per-token-latency as the context length exceeds 40K. Moreover, when increasing the KV cache size, the average memory usage of Inf-MLLM is always lower (around 13.5GB) than that of H2O (around 13.7GB) and StreamingLLM (13.7GB).
% the memory of Inf-MLLM grows linearly with the increase of cache size, and the per-token-latency remains stable as the context length exceeds 40K. Instead, abnormal results of latency are seen in H2O, which signify that its algorithm brings considerable additional cost. Thus, Inf-MLLM achieves up to 2x speedup compared with H2O, and maintains minimum memory usage against baselines.

\subsection{Demo of Streaming Inference On Edge}

We deploy Inf-MLLM on a Nvidia Orin GPU. Our method conduct long-term video stream understanding and multi-round QA continuously. As shown in Figure \ref{demo}, without our method, the vanilla Chat-Univi quickly reaches 25GB memory usage at Round 11 which keeps blowing up. On the other hand, with our method, the memory usage can be constrained, while the long term understanding capability is maintained (the model can reason about the very beginning of the video even at Round 11).

\subsection{Effects of Attention Bias}

% \subsection{Ablation Study}

% \begin{table}
% \centering
% \setlength{\tabcolsep}{1.9mm}
% \small
% \caption{Effects of attention saddles. 
% % To avoid the influence of attention bias, we default to set the bias parameter to 0.00001. 
% The perplexity is evaluated on the Wiki-Text-103 dataset.}
% \begin{tabular}{lccccc}
% \Xhline{1.2pt}
% \textbf{Attention Saddles} & 0\%   & 20\%  & 40\%  & 60\%  & 80\%  \\ \Xhline{1.2pt}
% Pythia-2.8B       & 12.20 & 12.06 & 12.02 & 11.99 & \textbf{11.99} \\
% LLaMA-2-7B-32K    & 10.80 & 8.90  & 8.55  & 8.41  & \textbf{8.35}  \\
% Vicuna-7B         & \textbf{11.08} & 11.75 & 11.72 & 11.68 & 11.63 \\ \Xhline{1.2pt}
% \end{tabular}
% \label{eval_saddles}
% \end{table}

% \subsubsection{KV Cache Configuration} In Table \ref{eval_saddles}, we ablate the effect by varying the percentage of attention saddles and recent tokens while maintaining the cache size as 2K. We can see that increasing the percentage of attention saddles can benefit perplexity for Pythia-2.8B and LLaMA-2-7B-32K. 
% The results validate the effects of attention saddles. 
% For Vicuna-7B, the best perplexity is achieved at 0\% for which only recent tokens are cached. This is because the model is only required to generate one next token based on the given text and does not need the long-term memory. 
% In the next part, we demonstrate how to enable models to capture the long-term dependency with the attention bias.
% In the next part, we prove that with attention bias, Inf-MLLM can further decrease the perplexity.

We evaluate the effects of attention bias in Table \ref{eval_bias}.  The experimental setup is similar to Section \ref{sec:long-term mem}. To capture longer-term dependency, smaller attention bias is required to reserve more former tokens and maintain long-term information. Table \ref{eval_bias} shows that as token distance scales up, the best value of attention bias decreases. However, when attention bias is smaller than 0.01, the accuracy rate drops to nearly zero due to model collapse on the long context. Therefore, it's essential to choose proper attention bias.

\begin{table}[thp]
\centering
\setlength{\tabcolsep}{1mm}
\small
\caption{Effects of attention bias on long-term memory. We evaluate it on the Vicuna-7B with KV cache size as 2K, and vary the token distance to evaluate the accuracy.}
\begin{tabular}{cccccc}
\Xhline{1.2pt}
\multirow{2}{*}{\textbf{Line Distance}} & \multirow{2}{*}{\textbf{Token Distance}} & \multicolumn{4}{c}{\textbf{Attention Bias}}  \\ 
\cmidrule(lr{1pt}){3-6}
                                                                                  & & 1             & 0.1           & 0.01 & 0.001 \\ \Xhline{1.2pt}
5 & 115                                                                                & \textbf{0.98} & 0.20          & 0.07 & 0.08  \\
15 & 345                                                                                & 0.66          & \textbf{0.97} & 0.07 & 0.08  \\
25 & 575                                                                                & 0.70          & \textbf{0.73} & 0.07 & 0.07  \\
35 & 805                                                                                & 0.48          & \textbf{0.90} & 0.06 & 0.06 \\
\Xhline{1.2pt}
\end{tabular}
\label{eval_bias}
% \vspace{-0.4cm}
\end{table}

% We evaluate Inf-MLLM separately on two LLMs, Vicuna and Pythia, and 

% In this section, we evaluate Inf-MLLM with Vicuna-7B \cite{vicuna}, Pythia-2.8B \cite{biderman2023pythia} and Chat-UniVi-7B \cite{chat_univi} on a single NVIDIA RTX4090 GPU . These models employ the RoPE position encoding for sequence generalization. Vicuna-7B and Chat-UniVi-7B are pre-trained with 4K lengths while the Pythia-2.8B is pre-trained with 2K lengths. We benchmark our Inf-MLLM on different video and text datasets and compare its performance against established baselines such as dense attention, window attention, and StreamingLLM.

\section{Conclusion}

Streaming inference of MLLMs encounters many challenges involving the under-performance on extended context and extensive memory consumption. The problem is more severe to deploy MLLMs on resource-constrained hardware like edge devices. In this paper, we observe attention saddles existing in attention maps of MLLMs, and introduce Inf-MLLM, an efficient framework to facilitate MLLMs to continuously handle long text and video streams on a single GPU without fine-tuning. Inf-MLLM contains an effective KV cache eviction mechanism to remove KV states of irrelevant tokens while maintaining a small size of KV cache during streaming inference. An adjustment strategy based on attention bias is proposed to further adjust the distribution of attention scores and avoid the accumulation in earlier tokens.
Experiments show that Inf-MLLM extensively extend the context length of MLLMs with texts up to 4 million tokens and 1-hour-long videos. 
% Inf-MLLM performs longer context dependency length and better ability to follow instructions than existing methods. 
% Inf-MLLM can efficiently model texts of up to 4 million tokens and 1-hour-long videos.

% \bibliography{refs}

\bibliography{aaai25}

\end{document}